\documentclass[twoside,11pt]{article}
\usepackage{jair, theapa, rawfonts}
\usepackage{graphicx}
\usepackage{xcolor}

\jairheading{Vol 76}{2023}{613-644}{01/2023}{03/2023}
\ShortHeadings{Liability in the Age of AI: a Use-Case Analysis of the Burden of Proof}
{Fernández Llorca, Charisi, Hamon, Sánchez \& Gómez}
\firstpageno{613}

\begin{document}

\title{Liability Regimes in the Age of AI: a Use-Case Driven Analysis of the Burden of Proof}

\author{\name David Fernández Llorca \email david.fernandez-llorca@ec.europa.eu \\
       \addr European Commission, Joint Research Centre (JRC)\\ Sevilla, 41092, Spain\\
       \addr University of Alcalá, Alcalá de Henares, 28805, Spain\\
       \AND
       \name Vicky Charisi \email vasiliki.charisi@ec.europa.eu \\
       \addr European Commission, Joint Research Centre (JRC)\\ Sevilla, 41092, Spain\\
       \AND
       \name Ronan Hamon \email ronan.hamon@ec.europa.eu \\
       \addr European Commission, Joint Research Centre (JRC)\\ Ispra, 21027, Italy\\
       \AND
       \name Ignacio Sánchez \email ignacio.sanchez@ec.europa.eu \\
       \addr European Commission, Joint Research Centre (JRC)\\ Ispra, 21027, Italy\\
       \AND
       \name Emilia Gómez \email emilia.gomez-gutierrez@ec.europa.eu \\
       \addr European Commission, Joint Research Centre (JRC)\\ Sevilla, 41092, Spain}


\maketitle

\begin{abstract}
New emerging technologies powered by Artificial Intelligence (AI) have the potential to disruptively transform our societies for the better. In particular, data-driven learning approaches (i.e., Machine Learning (ML)) have been a true revolution in the advancement of multiple technologies in various application domains. But at the same time there is growing concern about certain intrinsic characteristics of these methodologies that carry potential risks to both safety and fundamental rights. Although there are mechanisms in the adoption process to minimize these risks (e.g., safety regulations), these do not exclude the possibility of harm occurring, and if this happens, victims should be able to seek compensation. Liability regimes will therefore play a key role in ensuring basic protection for victims using or interacting with these systems. However, the same characteristics that make AI systems inherently risky, such as lack of causality, opacity, unpredictability or their self and continuous learning capabilities, may lead to considerable difficulties when it comes to proving causation. This paper presents three case studies, as well as the methodology to reach them, that illustrate these difficulties. Specifically, we address the cases of cleaning robots, delivery drones and robots in education. The outcome of the proposed analysis suggests the need to revise liability regimes to alleviate the burden of proof on victims in cases involving AI technologies.
\end{abstract}

\section{Introduction}
\label{Introduction}

AI-based systems have the potential to bring some benefits and opportunities to our societies, generating and transforming new products and services in multiple application domains. In recent years, there has been a real technological revolution in the advancement of AI, and more specifically, in ML and Deep Learning (DL) methods. Unlike conventional computer algorithms, in which the programmer explicitly implements the instructions needed to solve a particular problem, ML/DL approaches \footnote{From now on, the term AI will be used generically to refer to both ML and DL.} are based on the idea that the solution can be obtained by accessing data sufficiently representative of the problem and using a training procedure that allows fitting a mathematical model to such data. These approaches have been greatly benefited by the availability of massive data and improved computing power, which has enabled the use of increasingly complex models capable of solving increasingly sophisticated problems. In addition, this approach allows for continuous improvement of the system throughout its life cycle as more data and experience of use and interaction becomes available, making it a very powerful methodology. 

However, as this technology advances, there is growing concern about the risks to safety and fundamental rights that its adoption may entail \cite{whitepaperAI}. These risks arise mainly from certain intrinsic characteristics of certain AI approaches, such as lack of causality, opacity, unpredictability or the uncertainty derived from self and continuous learning capabilities. And although there are legal mechanisms proposed to minimize these risks, such as ex-ante requirements in the recently proposed European regulatory framework for AI \cite{AIAct}, general safety standards, or ethical and socially responsible AI approaches (\citeR{Cheng2021}; \citeR{Kim2021}), they do not exclude the possibility of some end user or bystander being harmed. In such cases, victims should be able to seek compensation, and the most typical way to do so is on the basis of liability regimes, in particular tort law (e.g., see Chapter 8, Parts 5 and 6 by \citeR{Koziol2015}). In addition, recent advances in AI allow the degree of human intervention and supervision to become less and less, which also brings with it the need to clarify the attribution of responsibility. 

Generally speaking, we can identify three legal frameworks by which victims can obtain compensation for product-induced damages \cite{EGEC2019}:

\begin{itemize}
\item \textbf{Fault-based liability}: where the injured parties or claimants have to prove, in principle, that the defendant or wrongdoer (or its employees in the case of vicarious liability) caused the damage intentionally or negligently. This involves identifying the applicable \emph{standard of care} the defendant should have fulfilled, and proving that it was not fulfilled. In the language of negligence the issue would be whether the product or some of its key components were negligently designed, manufactured, deployed, secured, maintained, updated, monitored, marketed, operated, used, etc. 

\item \textbf{Strict liability} (a.k.a. risk-based liability): based on the understanding that someone is allowed to use a dangerous object or perform a risky activity for her own purposes, so this person also bears the loss if such risk materialises \cite{Karner2021}. Therefore the victim does not need to prove the misconduct on the part of any wrongdoer. Instead, the injured parties only have to prove that the risk arising from the sphere of the liable party actually materialised (i.e., the risk subject to strict liability materialises). 

\item \textbf{Product liability}\footnote{We focus primarily on product liability at the EU level, which a subcategory of strict liability but based on defectiveness. Note that although in most US states product liability is linked to a defect, this is not applicable in all cases.}: where victims can claim against the producer (manufacturer, importer, supplier or seller) for a defect present at the time the product was placed on the market. Injured parties or claimants have to prove that the product was defective (irrespective of fault) and the causal link between the defect and the damage. A defective product is a product with a (usually unintended) flaw, i.e., the finished product does not conform to the producer’s own specifications or requirements or to general safety requirements. In this sense, the product fails to provide the safety that the public at large is entitled to expect (i.e., \emph{standard of safety}), and this lack of safety causes the damage. In the language of defectiveness (similar to fault) the product or some of its key components may have design, manufacturing, maintenance or marketing defects \cite{Vladeck2014}. 
\end{itemize}

The same features that make AI systems inherently risky, such as lack of causality, opacity, unpredictability or their self and continuous learning capabilities, may lead to considerable difficulties when it comes to proving defect or fault and causal link with the damage in the different liability regimes. In this paper, we illustrate these difficulties presenting three case studies, as well as the methodology to reach them. Specifically, we address the cases of cleaning robots, delivery drones and robots in education.

\section{Related Work}

The impact and challenges that AI systems have on liability rules have been extensively studied over the past few years. One of the first points of discussion focuses on the legal personality that could be attributed to an autonomous AI-based system that makes decisions without human intervention or supervision. As highlighted by \citeA{Cerka2015}, neither national nor international law recognizes AI as a subject of law, which means that an AI system cannot be held personally liable for the damage it causes. The question to be asked would be who is liable for damages caused by the actions of AI. On the one hand, when the AI system simply acts  as a tool to provide humans with additional knowledge, recommendations, etc., the person operating that tool would then be ultimately responsible for any decisions (e.g., the learned intermediary doctrine, \citeR{Sullivan2019}). On the other hand, as AI becomes more autonomous, it will be more difficult to determine who or what is making decisions and taking actions \cite{Shook2018}. Attributing legal personality to AI systems may be a possibility in some cases, but as concluded by the Expert Group on Liability and New Technologies set up by the European Commission for the purposes of liability \cite{EGEC2019}, it is not necessary to give autonomous systems a legal personality, as the harm AI systems may cause can and should be attributable to existing persons or organizations. Harm caused by AI systems, even fully autonomous ones, is generally reducible to risks attributable to natural persons or existing categories of legal persons, and where this is not the case, new laws directed at individuals are a better response than the creation of a new category of legal person \cite{Abbott2019}.

Another point of debate has been to establish which \emph{liability framework} is the most appropriate for AI systems. One of the first questions was to decide whether to apply a standard of care (negligence-based) or a standard of safety (defect-based). As discussed by \citeA{Vladeck2014}, and unlikely as it may seem, this issue 
was addressed by a court over six decades ago (in 1957, based on US tort law) concerning a car accident due to one pedestrian (the claimant) crossing inappropriately. In its analysis, the court argued that the driver, however efficient, is not a mechanical robot capable of avoiding an accident in such circumstances. Implicitly, the court argued that an autonomous vehicle must operate according to a safety standard. As explained further by \citeA{Vladeck2014}, in modern product liability law, such a standard would likely be set on a risk-utility basis (strict liability). In the past, claims related to product failures were initially handled under negligence theories, largely because the focus was on the conduct of humans, not the performance of machines. However, as negligence claims related to product failures became more difficult to prove, strict liability principles took root to govern product liability cases, especially in the case of complex and highly autonomous products such as modern AI systems. 

As suggested by \citeA{Giuffrida2019}, liability for a defective product applies when, among other possibilities, a reasonable alternative design (or manufacturing, maintenance or marketing processes) could have avoided or limited foreseeable risks of harm\footnote{This approach corresponds mainly to the US doctrine, which can be seen as a possible path within the broader EU approach, where defect is primarily linked to the safety that the general public expects from a product.}. For complex AI systems, deciding who is responsible and for what when someone has been injured can be extremely complicated as many parties come into play in the complex supply chain. For example, there are, among others: AI developers, developers of training frameworks, data collectors, 
annotators, controllers and processors, providers of  
AI systems integrated in other AI systems, manufacturers of the products incorporating the AI systems, users of these products, etc. On top of that, we find the aforementioned specific characteristics of AI systems (i.e., lack of causality, opacity, unpredictability and self and continuous learning) which makes it substantially more difficult to demonstrate causal relationships \cite{Bathaee2018}. Therefore, the burden of proving a design alternative that might have avoided harms can be huge and, in some cases, unfeasible for the victim to address. As explained by \citeA{Buiten2019} with respect to arguing and evaluating how complex algorithms (e.g., AI systems) may have caused harm, it is obvious that courts and injured parties remain at a disadvantage compared to the expert developers producing the systems. 

The problem of the difficulty in proving causation for AI systems has been clearly identified in the literature, whether for fault-based or defect-based liability regimes. As described by the Expert Group on Liability and New Technologies, regarding wrongfulness and fault: “\emph{In the case of AI, examining the process leading to a specific result (how the input data led to the output data) may be difficult, very time-consuming and expensive}”. And with respect to product liability: "\emph{the complexity and the opacity of emerging digital technologies complicate chances for the victim to discover and prove the defect and prove causation}" \cite{EGEC2019}. In view of these difficulties, experts and academics propose various alternatives to alleviate the burden of proof on victims, including the reversal of the burden of proof, rebuttable presumptions, or the application of strict liability regimes, among others (\citeR{Karner2021}; \citeR{Wendehorst2021}). From a technical perspective, the need of explainable AI has been also proposed as a mean to overcome the AI opacity issue (\citeR{Padovan2022}; \citeR{Fraser2022}). 

In this paper, we focus on causation in the context of AI and liability to present a set of use cases, including legally relevant technical details, which illustrate the specific difficulties involved in AI systems when it comes to proving causation in liability regimes, either from a standard of care or from a standard of safety point of view. Contrary to previous works (\citeR{EGEC2019}; \citeR{Karner2021}; \citeR{Erdelyi2021}) the objective of our use case analysis is not only to test the legal or insurance issues, but to address the technical difficulties that even in the best case scenario, i. e., with the cooperation of the defendant who provides access to documentation, training data and logs \cite{Wendehorst2021}, an expert would have to face in order to prove fault or defect. 
The same specific AI challenges related to the burden of proof identified in the context of anti-discrimination law (\citeR{Hacker2018}; \citeR{gerards2021}) also apply to liability regimes. In this paper, we provide a more specific analysis including use cases at a high level of granularity.
We mainly focus on systems that can produce physical damage, so the link between defect or fault and the harm should be also established. We therefore omit pure software based cases such as the recent Australian case of ACCC v Trivago \cite{Fraser2022}. The selected examples represent recent technological developments, potentially available within a relatively short time horizon, which may pose risks to third parties. In this analysis, we link the difficulties in addressing the burden of proof to the characteristics of certain AI systems such as lack of causality, opacity and unpredictability. 

The structure of the rest of the paper is as follows. First, we describe those specific features of AI that we consider pose a problem in demonstrating causality when attributing responsibility in liability regimes. Second, we present the methodology for generating the use cases. Third, we describe the use cases and, finally, provide a general discussion, conclusions and future work.

\section{Specific Features of Certain AI Systems}\label{features}

Unfortunately, there is no commonly accepted technical definition of AI that is valid for multiple contexts \shortcite{Samoili2021}.  Generally speaking, we can say that AI is a broad concept related to machines capable of making decisions and performing tasks in a way that we would consider intelligent or human-like. However, although AI is often discussed in general terms, 
most of the features linked to safety and human rights risks are mainly present in recent ML/DL approaches.
Indeed, in these cases the methodology changes substantially from conventional computational systems. Instead of explicitly implementing the instructions needed to address a particular problem, data-driven AI techniques capture data sufficiently representative of the problem to be solved, and fit a complex mathematical model with many parameters that are determined during a training process. This approach has been conveniently reinforced by the growing availability of massive datasets, as well as increasingly powerful computational systems. They have allowed the use of more and more sophisticated mathematical models with a greater number of parameters capable of dealing with increasingly complex problems. Moreover, this approach allows for continuous training of the system throughout its life cycle as more data and experience of use and interaction becomes available. This enables continuous improvement without a substantial change in methodology. 

On the one hand, the \textbf{complexity} and power of AI systems allows to implement solutions that perform tasks and make decisions with increasing \textbf{autonomy}\footnote{Note that some research communities prefer to use "automation" rather than "autonomy". The trend we observe is that both concepts can be used interchangeably, although autonomy can also contemplate actions beyond the intended domain.}, understood as \emph{the capability to perform with an absent or low degree of external influence or human involvement} \shortcite{glossary22}. 
However, it is important to note that complexity and autonomy are not exclusive characteristics of AI systems. Complexity is a broader concept related to the whole system (number of interconnected elements, complexity of the supply chain, etc.) and to the operating environment (number and type of agents, interactions, structure of scenarios, etc.). The level of autonomy, although dependent on the capacity of the system, is still a design variable.

On the other hand, these approaches may suffer from multiple limitations that need to be properly addressed. These include potential \textbf{bias} in the data that is indirectly incorporated into the trained model \cite{Songul2018}\footnote{Bias in AI can be more broadly defined as \emph{an anomaly in the output of AI systems, due to the prejudices and/or erroneous assumptions made during the system development process or prejudices in the training data, so the results from the AI system cannot be generalised widely} \cite{glossary22}.}, possible problems of \textbf{overfitting} \cite{overfi2019}, i.e., an excessive fit to the data that leads to a \textbf{lack of generalisation}\footnote{\emph{Generalisation} refers to the ability of the model to adapt adequately to new, previously unseen data, drawn from the same distribution as that used to create the model.}, or the problem known as the \textbf{curse of dimensionality}\footnote{The amount of data needed to represent a problem grows exponentially with the dimension of the input space (i.e. with the number of variables or features).} \cite{curse2005}, especially relevant in problems with multiple inputs. But again, it is important to note that bias issues, or the ability to find a complete or sufficiently general solution to a problem, or the difficulties in finding solutions to problems in high dimensional spaces, are not unique to AI systems. Bias is introduced in both algorithm design and data, but it is a feature of virtually all designs carried out by humans and, of course, is an intrinsic feature of human decision-making. The dimensionality problem is an extrinsic feature that refers to the complexity of the operating scenario and the number of variables involved in the system output. This is also related to the need for data, as the number of data samples required for an acceptable representation of the input space increases exponentially with dimensionality. Finally, overfitting or lack of generalisation, while tending to be more prominent when systems are more complex and flexible (e.g. ML/DL), other more conventional computational systems (e.g. expert systems or a set of fixed rules) can suffer from the same problem, for example, due to bad specification or poor design.

\begin{figure}[ht]%
\centering
\includegraphics[width=0.9\textwidth]{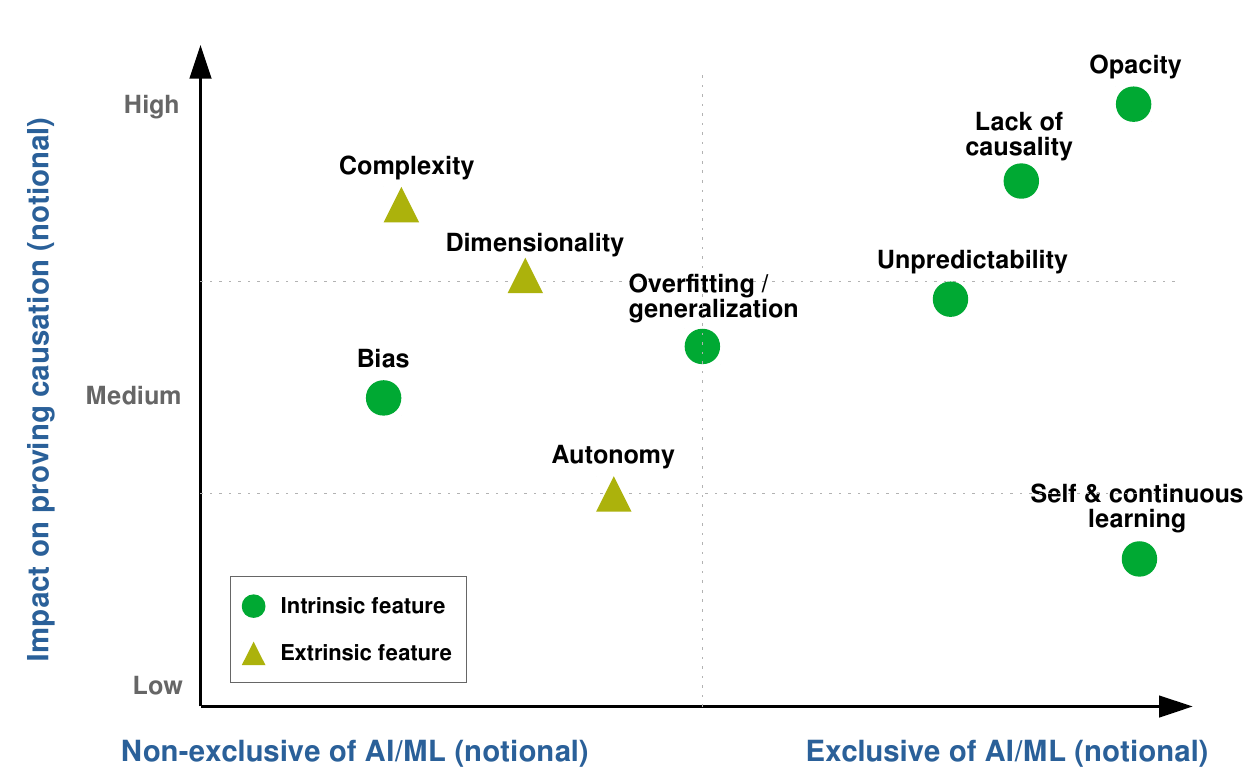}
\caption{Notional representation of the intrinsic and extrinsic features of AI 
systems and their impact on proving causation. The abscissa represents the uniqueness of each feature of AI systems compared to those based on more conventional computational methods.}\label{fig:features}
\end{figure}

In our view there are four specific characteristics of certain AI systems that could pose significant challenges to proving causality in liability regimes, namely, \textbf{lack of causality}, \textbf{opacity}, \textbf{unpredictability} and \textbf{self and continuous learning capability}
. These features are described in detail below. In Fig. \ref{fig:features} we illustrate all the aforementioned characteristics of AI  systems and their impact on the difficulty when proving causation. As can be seen, their impact and degree of exclusivity in relation to AI systems is conceived more as a continuum than as a discrete categorisation. 

\subsection{Lack of Causality}
Causality can be defined as a generic relationship between two variables: the effect and the cause that gives rise to it. As described by \shortciteA{Guo2021}, when learning causality with data, we need to be aware of the differences between statistical associations and causation. The current success of data-driven AI is mainly based on its ability to find correlations and statistical patterns in large-scale, high-dimensional data (i.e., statistical modelling). And not just any type of data, but usually \emph{independent and identically distributed (i.i.d)} data. That is, random observations that are not dependent on each other and have the same probability of occurring (e.g, rolling a die several times). Problems can be intrinsically i.i.d, or they can be made approximately i.i.d. and, in such cases, convergence of a learning algorithm can be theoretically guaranteed at the lowest achievable risk \shortcite{Scholkopf2021}. Therefore, it is not surprising that, with sufficient data, data-driven AI can surpass human performance. 

The i.i.d. assumption can be violated in the “independent” part and/or in the “identically distributed” part. On the one hand, if there are statistical dependencies between variables, this assumption does not allow explicit modelling and learning of the causal relationships between them. On the other hand, if the distribution from which the data is derived differs slightly between the training phase and real-world operation, the data-driven AI system often performs poorly. For example, computer vision systems can fail recognizing objects when they appear under new lighting conditions, different viewpoints, or against new backgrounds \shortcite{Barbu2019}. 

In most cases, real-world data do not fully satisfy the i.i.d. hypothesis. As described by \citeA{Peters2017} and \citeA{Scholkopf2021}, \emph{generalising well outside the i.i.d. setting requires learning not mere statistical associations between variables, but an underlying causal model}. Learning causal relations is highly complex, and it requires collecting data from multiple domains (i.e., multiple distributions) as well as the ability to perform interventions (i.e., interventional data) that trigger a change in the data distribution \cite{Scholkopf2021}. 

Causality (or lack thereof) is highly correlated with the ability of data-driven AI systems to respond to unseen situations (linked to unpredictability and generalisation capabilities) and to remain robust when some interventions change the statistical distribution of the target task. This includes \textbf{adversarial attacks}, where carefully selected perturbations of the inputs, typically imperceptible or inconspicuous to humans (e.g., adding invisible noise to images, or inverting letters in a text) can induce large variations in the system's outputs \shortcite{Papernot2016} (constituting violations of the i.i.d. assumption). In addition, learning causality can help ML systems to better adapt to other domains (e.g., multi-task and continuous learning), and improve interpretability and explainability of ML systems \shortcite{Chou2022} (linked to opacity as presented in Section \ref{subsec:opacity}). It has also been shown that explainable AI focused on causal understanding ("causability") can support effective human AI interaction \cite{Holzinger2021}.

However, despite numerous advances in this field (\citeR{Pearl2009}; \citeR{Peters2017}), learning causal relationships still poses numerous challenges and so far, according to \citeA{Scholkopf2021}, data-driven AI \emph{has neglected a full integration of causality}. But in any case, forcing AI systems to rely on causal mechanisms rather than correlations will not ensure “intuitive” models, and we have to assume that certain phenomena may not be exhaustively based on causal mechanisms \cite{Selbst2018}. 

\subsection{Opacity}
\label{subsec:opacity}

The formal definition of opacity refers to obscurity of meaning, resistance to interpretation or difficulty of understanding. In the AI domain it is also known as the \textbf{black-box} effect \cite{Castelvecchi2016} since the decision-making process with ML appears inscrutable from the outside. Even when AI experts, or the creators of the AI system themselves, examine the system (source code of the model and the training process, the model architecture, the trained parameters, the training, validation and test datasets, etc.), it is difficult or impossible to understand how they combine to form a decision. Opacity arises from the inability to provide human-scale reasoning from complex AI models \cite{Burrell2016}. Another similar term for the fact that the rules governing the AI system are so complex, numerous, and interdependent that they frustrate human comprehension is "inscrutability" \cite{Selbst2018}.

This intrinsic feature of certain AI systems has prompted the development of \textbf{transparency} requirements, whether horizontal (e.g., the AI Act \citeR{AIAct}) or sector-specific (e.g., possible requirements in the field of autonomous vehicles as described by \citeA{Llorca2021}). The compliance with transparency requirements (which include measures to address \textbf{traceability}\footnote{Traceability of an AI system refers to the \emph{capability to keep track of the
processes, typically by means of documented recorded identification}, or to the \emph{ability to track the journey of a data input through all stages} \cite{glossary22}.}, \textbf{interpretability}
\footnote{AI models are interpretable \emph{when humans can readily understand the reasoning behind predictions and decisions made by the model} \cite{glossary22}.}, or \textbf{explainability}\footnote{Explainability in AI can be understood as a \emph{feature of an AI system that is intelligible to non-experts}, or as \emph{methods and techniques in AI such that the results of the solution can be understood by humans} \cite{glossary22}}) will alleviate the burden of proving causality. However, the well-known trade-off between accuracy and interpretability in AI systems remains as an obstacle, i.e., more accurate models are less interpretable and vice versa. Furthermore, attempts to explain black-box machine learning models may not be sufficient to demonstrate causality \cite{Rudin2019}. Moreover, the field of explainable AI still needs stronger conceptual foundations \shortcite{Cabitza2023} and to overcome limitations regarding stability, robustness and comprehensibility of explanations \cite{saarela2022robustness}. Therefore, despite efforts to develop interpretable systems by design or implement post-hoc explanations, the problem of opacity of complex AI systems is likely to remain one of the most critical issues for victims when trying to prove causation in either fault or defect schemes. 

\subsection{Unpredictability}
Although unpredictability has often been associated with the self and continuous learning capability of AI systems, this characteristic is also intrinsic to static or "frozen" systems,  i.e. systems that do not continue to learn and adapt while in operation.

\begin{figure}[ht]%
\centering
\includegraphics[width=\textwidth]{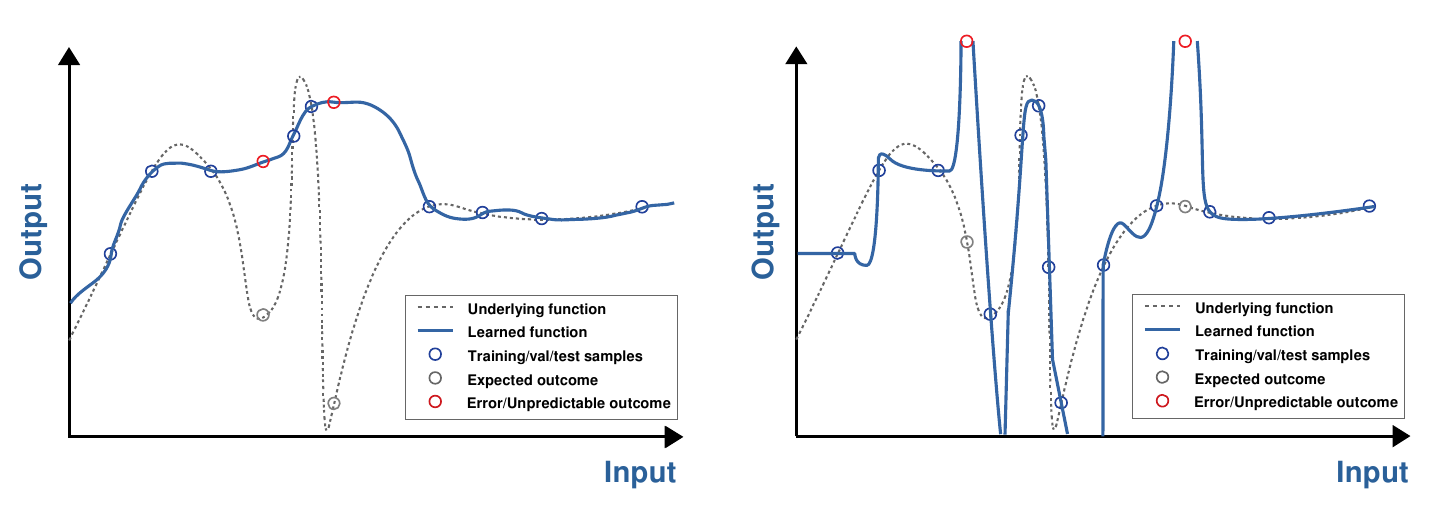}
\caption{Illustration of unpredictability issues due to (left) poor representation of the input data space and (right) overfitting. The underlying function represent the desired, unknown pattern. Two examples of erroneous and unpredictable results (red circles) compared with the expected outcome (gray circles) are depicted on each case. This is a simple scenario with only one input. Normally the dimension of the input data space is much larger. }\label{fig:unpred}
\end{figure}

Unpredictability in data-driven AI systems is mainly due to two reasons. First, it can apply in the case where the \textbf{dataset is not sufficiently representative} of the problem to be addressed by the machine learning model. Regardless of the generalisation capability of the model and the training process, the solution found in the poorly represented regions of the input space will generate unpredictable results. We illustrate this effect in Fig. \ref{fig:unpred} (left). As can be seen, if the underlying (unknown) pattern of what is to be learned is very complex and the training, validation and test data are not sufficiently representative, the function learned by the data-driven AI model may not fit the real nature of the problem and generate totally unpredictable values (which may lead to safety issues depending on the scenario). This problem is very significant in cases of high dimensionality and complexity of the input space, the operating environment, the number and type of interactions with agents, etc., where obtaining sufficiently representative data is a major challenge. 

The second reason refers to the \textbf{lack of generalisation or overfitting} of the learned function. Even in the case where the input space is reasonably well represented by the data set (which, after all, is always a limited sample of the entire input space), when the learning process over-fits the data, the outcome of the learned function for samples not available during training, validation and testing can be totally unpredictable. This effect is illustrated in Fig. \ref{fig:unpred} (right).

In both cases, these effects imply that even small, almost imperceptible, linear changes in inputs can produce abrupt and unpredictable changes in outputs. This becomes more pronounced the larger the non-linear relationship between inputs and outputs. Moreover, it is these constraints that allow exploiting the aforementioned \textbf{adversarial attacks}. 

In addition, it is important to highlight certain types of AI systems, that is, \textbf{recurrent models} \cite{Mandic2001}, in which the output not only depends on the inputs but also on the internal state of the model. This means that the same input at two different time instants (with two different model states) may provide different outputs. This feature allows dealing with dynamic problems where states and inputs prior to the instant of decision are relevant. But this feature also implies that, given the same inputs, small variations in the state of the model can generate different results, which implies a source of unpredictability. 

Finally, it should be noted that unpredictability can lead to significant problems of \textbf{repeatability}\footnote{\emph{The measurement can be obtained with stated precision by the same team using
the same measurement procedure, the same measuring system, under the same operating conditions, in the same location
on multiple trials} \cite{glossary22}.}, \textbf{reproducibility}\footnote{\emph{The measurement can be obtained with stated precision by a different team and a different measuring system, in a different location on multiple trials} \cite{glossary22}.} or \textbf{replicability}\footnote{\emph{The measurement can be obtained with stated precision by a different team
using the same measurement procedure and the same measuring system, under the same operating conditions, in the
same or a different location on multiple trials} \cite{glossary22}}. While conventional computational approaches are less likely to provide different results under the same operating conditions, the unpredictability issues described above pose an additional difficulty in ensuring system consistency and robustness. 

\subsection{Self and Continuous Learning}
Although the terms \textbf{self-learning} and \textbf{continuous learning} have often been used in a somewhat vernacular way, for example by mixing them, or by referring to only one of them with both meanings, these are two distinct terms with different meanings. 

On the one hand, \textbf{self-learning} refers to the ability of the AI system to \emph{recognize patterns in the training data in an autonomous way, without the need for supervision} \cite{glossary22}. This definition implicitly refers to \emph{unsupervised learning} (i.e., learning that makes use of unlabelled data during training \shortcite{glossary22}) but explicitly stating that this is done without human supervision (i.e., autonomously). Nevertheless, this term does not specify if the process is performed off-line or while the system is in operation. Another definition refers to \emph{learning from an internal knowledge base, or from new input data, without introduction of explicit external knowledge} \shortcite{glossary22}. This definition implicitly mentions \emph{domain adaptation} and might suggest that the process is autonomous.

On the other hand, \textbf{continuous learning} refers to \emph{incremental training of an AI system that takes place on an ongoing basis during the operation phase of the AI system life cycle} \cite{glossary22}. Therefore, this term explicitly excludes off-line learning as it specifically refers to on-line learning while in operation. Although the level of human supervision is not included in the definition, we can implicitly assume that if the operation of the system is autonomous, continuous learning also takes place autonomously. The aforementioned characteristics of lack of causality and unpredictability make this approach particularly risky in certain scenarios. Additionally, it is worth mentioning an effect known as \textbf{catastrophic forgetting}. That is, under certain conditions, the process of learning a new set of patterns (in this case continuously during operation) suddenly and completely interferes, or even erases, the model’s knowledge of what it has already learned \cite{French1999}. 

Self and continuous learning are crucial for systems operating in changing environments, as they enable the acquisition, fine-tuning, adaptation and transfer of increasingly complex knowledge representations. For example, they are widely used in contexts where adaptation to the profile of each user is required. They are also commonly used during the development phase of complex AI systems (e.g., reinforcement learning). However, once systems are deployed, and in cases where they may cause damage to persons or property (triggering liability mechanisms), such approaches may involve unacceptable risks. As highlighted by \citeA{Wendehorst2021} when "AI is learning in the field" the system adapts its function and behaviour after the deployment and thus it is very difficult to assess the risk because its risk profile can significantly change over time. Moreover, systems (e.g., hardware, platforms) may also be damaged, with the costs that this entails. 

From the perspective of fault- and defect-based liability regimes, it is clear why self and continuous learning are a major challenge, as they are directly related to the question of foreseeability \cite{Rachum-Twaig2020}. It is reasonable to assume that a defendant will only be held liable if it could reasonably foresee and prevent the potential results of an action \cite{Benhamou2021}. If an AI system with self and continuous learning capabilities is placed on the market and causes harm to a person, it would be very difficult for the claimant to prove that the system was negligently designed or defective, and especially that these issues existed when the system left the hands of the providers \cite{Cerka2015}. That is one of the reasons for suggesting that the \textbf{development risk defence}, which allows the provider to avoid liability if the state of scientific and technical knowledge at the time when the product was put into circulation was not such as to enable the existence of the defect to be discovered, should not apply in these cases \cite{EGEC2019}.

Safety regulations, or specific AI regulations (e.g., the AI Act \citeR{AIAct}), are conceived to certify that a specific "version" of a system complies with established requirements. Enabling self and continuous learning is highly unlikely, as these features may substantially modify the behaviour of the system after certification. This could lead to cases where two AI systems of the same type, exposed to different environments, would differ in their behaviour over time. For example, in the context of vehicle safety regulations, there is preliminary consensus that self and continuous learning should be not be allowed, as they are incompatible with existing regulatory regimes and safety expectations \cite{UNECE2021}. 

Another example can be found in the framework of the AI Act. High-risk AI systems that “continue to learn” after being placed on the market or put into service shall be developed with appropriate mitigation measures. Whenever there is a change in the behaviour that may affect compliance with the regulation (i.e., a substantial modification) the AI system shall be subject to a new conformity assessment (Article 43(4) of the AI Act proposal \citeR{AIAct}). In other words, self and continuous learning are only allowed if the provider can predetermine and foresee the changes in the performance. Depending on the context, this may be virtually impossible. For example, we can consider cases where behavioural adaptation through self and continuous learning is necessary to improve the interaction with end-users, but only if the change in the AI system does not affect their safety. 

Accordingly, when we refer to the issue of lack of foreseeability or predictability of AI systems, and how it challenges different liability regimes, it is more realistic to mainly focus on the intrinsic features of lack of causality and unpredictability rather than on self and continuous learning. It is reasonable to assume that safety or AI specific regulations will not allow for the integration of self and continuous learning if they can affect the behaviour of the system after certification, especially in cases where AI systems may pose a risk of physical harm. That is the main reason why these characteristics appear with less impact in Fig \ref{fig:features}.

\section{Use Case Selection Criteria and Descriptive Structure}\label{methodology}

The main goal of this work is to develop a number of use cases to illustrate the specific difficulties of AI systems in proving causation, and to address the technical challenges that an expert would face in proving fault or defect, including legally relevant technical details. In this section we describe the inclusion criteria and the proposed structure for describing the use cases. 

\subsection{Inclusion Criteria}
\begin{figure}[ht]%
\centering
\includegraphics[width=\textwidth]{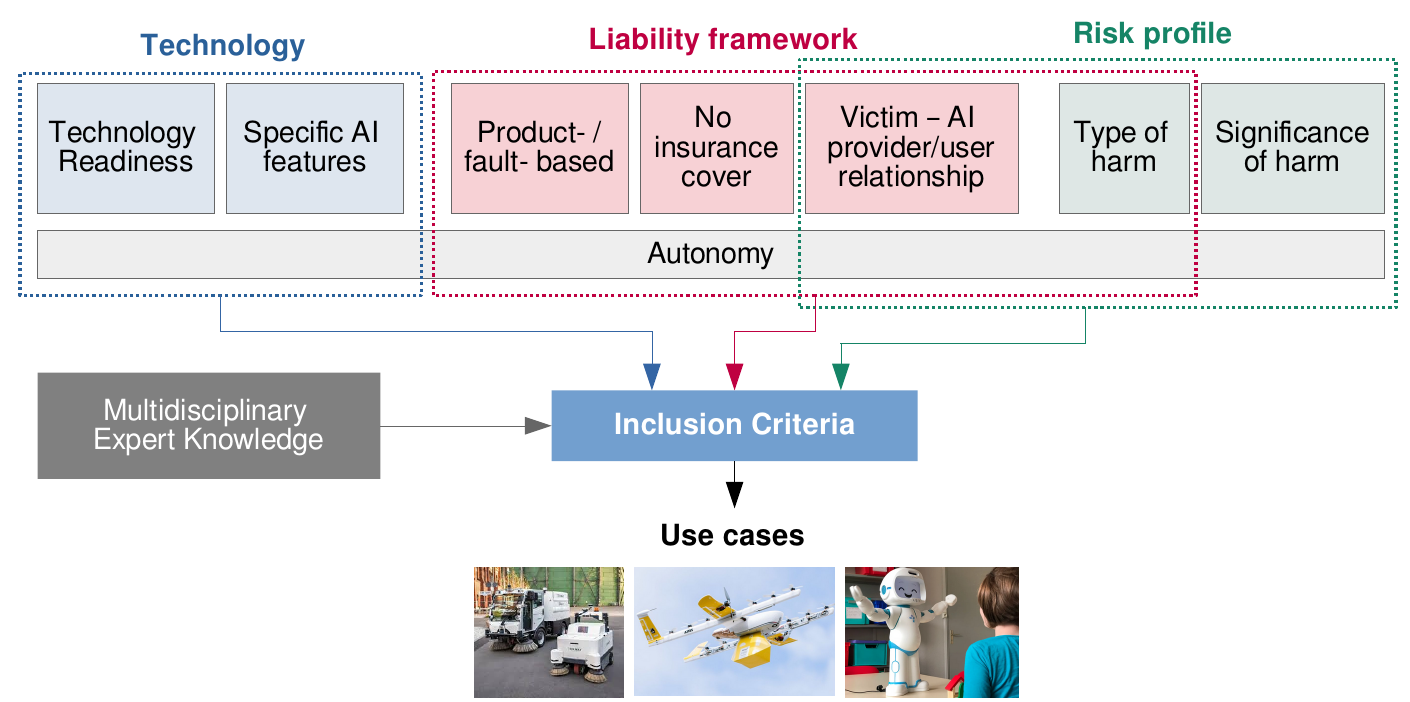}
\caption{A schematic view of the proposed methodology to develop the use cases.}\label{fig:method}
\end{figure}

For the identification and elaboration of the use cases we have developed a three-pillar inclusion criterion that comprises multiple elements related to the technology, the liability framework and the risk profile. An overview of the proposed methodology is shown in Fig. \ref{fig:method}. The proposed sets of criteria would need to be fulfilled cumulatively. In the following, the different criteria used for the development of the use cases are described. 

\subsubsection{Technology Aspects}
From a technology point of view, the proposed approach takes into account two main aspects.

\begin{itemize}
    \item \textbf{Technology readiness:} the use cases should include AI systems that are sufficiently realistic to be described with a sufficient degree of granularity and confidence. This could refer to experimental platforms that could have been used in relevant but restricted environments or piloted under certain conditions in operational environments.     Therefore, they do not necessarily have to be products currently on the market, but they should have a relatively high level of technological maturity and availability. For example, if we use the Technology Readiness Levels (TRLs) \shortcite{TRL2022}, we focus on systems between TRL 5 and 7. 
    \item \textbf{Specific AI features:} we focus on the specific features of certain types of AI described in Section \ref{features}, that is, lack of causality, unpredictability, opacity and, in some cases, self and continuous learning. As described above, we can assume that safety or AI specific regulations will not allow self and continuous learning if they can affect the behaviour of the AI system in a potentially dangerous way. This applies in particular to products subject to type-approval or ex-ante conformity assessment (usually products that may pose a risk to the physical harm to persons). 
\end{itemize}

\subsubsection{Features of the Liability Framework}
As far as the liability framework is concerned, we consider three main aspects: the liability regime, the lack of insurance and the type of relationship between the victim and the provider/manufacturer or user/operator. An additional criterion (the type of damage) can also be linked to this section, but is mainly associated with the general context. 

\begin{itemize}
    \item \textbf{Liability regimes:} the main frameworks considered are the product liability and fault-based liability regimes, based on proof of defect and fault, respectively. The idea, therefore, is to avoid scenarios where risk-based liability is directly applicable, or where it is at least compatible with product- and fault-based approaches. 
    \item \textbf{Lack of insurance:} given that the purpose is to identify situations where the victim must bear the burden of proof, it is more effective to avoid approaches where the damage is covered by insurance \cite{Erdelyi2021}. That is the case, for example, for autonomous vehicles, which will be covered by motor insurance schemes. This criterion is mainly related to the avoidance of risk-based liability, which is usually supported by mandatory insurance.   
    \item \textbf{Victim - AI provider/user relationship:} to avoid triggering consumer contractual law mechanisms that could prevail over liability approaches, the victim and the AI system provider or user should not have any contractual relationship. That is, scenarios should be based on the damage caused to third parties (e.g. bystanders). Therefore, applications intended to be used outside of private environments would be considered suitable for the use cases included.
\end{itemize}

\subsubsection{Risk Profile}
Once the type of relationship between the victim and the provider or user of the AI system has been established, the next step is to define the risk profile, depending on the type of potential harm and its significance.

\begin{itemize}
    \item \textbf{Type of harm:} since the AI-specific issues of proof of causation can be applied irrespective of the type of harm, our analysis focuses on any type of harm compensable under all liability regimes, i.e. personal injury, including medically recognised psychological damage, and property damage. In this way we ensure the broad relevance of the use cases. 
    \item \textbf{Significance of harm:} although the use cases may not necessarily put the "general public" at risk, the possibility of harm to third parties in the envisaged scenarios should be reasonably high. In other words, the scale of potential accidents should not be negligible. To realistically achieve this goal, on the one hand, we are looking for applications operating in public environments. On the other hand, to provide illustrative examples, we consider that AI systems should be embodied in some kind of mobile robotic platform, whose operating dynamics are more likely to pose a danger to bystanders.
\end{itemize}

\subsubsection{Autonomy}
As mentioned in Section \ref{features}, we refer to autonomy as the ability of the AI system to perform without external influence or human involvement. This feature cuts across the three sets of requirements defined for the development of the inclusion criterion. On the one hand, and although the design of autonomous systems is not unique to AI-based approaches, achieving high degrees of autonomy, without requiring human intervention or supervision, is possible today thanks to recent technological advances in AI. On the other hand, the fact that the system does not require human intervention allows liability regimes to be triggered not towards a person responsible for operation or supervision, but towards the AI provider or user. This way, it would be difficult to link a human action or omission to a certain output that caused the damage and the proof of causation would necessarily focus on the intrinsic behaviour of the AI system itself. Finally, the ability of the system to operate in public environments in a fully autonomous fashion is also related to the risk profile, the type of damage and its significance. 

\subsection{Expert Knowledge}
Followed by the specification of the inclusion criteria, we considered a number possible scenarios that would fulfil the aforementioned requirements. At this stage, as illustrated in Fig. \ref{fig:method}, the interaction among the members of our team with different disciplinary background (law, engineering, computer science, robotics, human cognition, human factors and social science) was pivotal for the final identification of the use cases.

\subsection{Use Cases Structure}

\begin{figure}[ht]%
\centering
\includegraphics[width=0.7\textwidth]{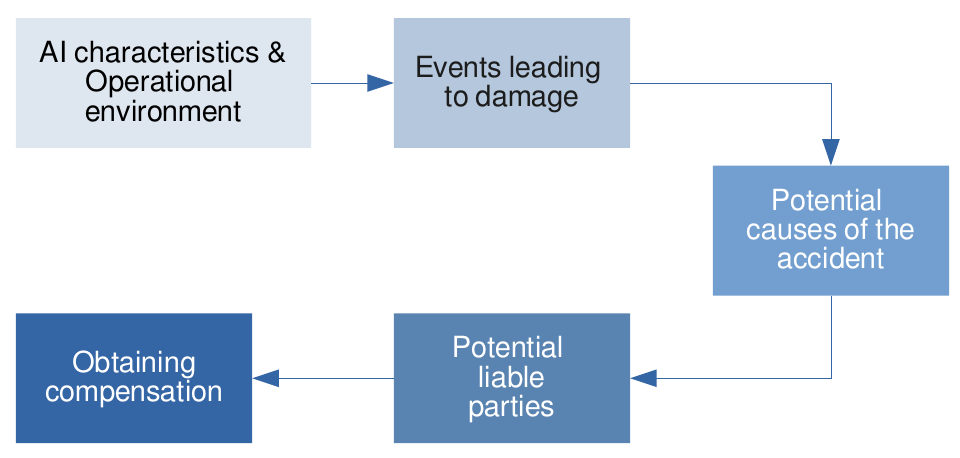}
\caption{Proposed structure to describe the use cases.}\label{fig:structure}
\end{figure}

As can be observed in Fig. \ref{fig:structure}, for each use case, we address the following steps. 
First, we describe the main characteristics of the AI system(s) involved in the application and the operating environment. Second, we describe the hypothetical events leading to the damage, as well as the damage itself. Third, we identify the possible causes of the accident. Fourth, we identify the possible liable parties. And finally, we detail the procedure for obtaining compensation, highlighting the requirements and the main difficulties faced by the victim. 

\section{Case Studies}\label{cases}

\subsection{General Considerations}
This work focuses on fault- and defect-based liability regimes. Although a strict liability regime could apply in some cases, where the claimant would only have to prove that the risk arising from the sphere of the liable party (i.e., the user or the operator) materialised, it is very likely that the user/operator would take recourse against the providers of the end product or the individual AI components. Therefore, the need to prove fault or defect could also apply.  

Under national tort law, the claimants would in principle have to prove that the defendant caused the damage negligently. That involves proving non-compliance with the applicable standard of care (i.e., fault) and the causal relationship with the damage. Under product liability, the claimants would have to prove that the AI-based product was defective and the causal link with the damage. 

In both cases, expert opinion, access to technical documentation on the design, functioning and operation of the system, as well as access to relevant data and system logs (e.g, inputs, outputs, internal states of the subsystems) corresponding to the last few minutes before and after the accident, would be required. The expert would have to understand in considerable detail the overall functioning of the system, components involved, inter-dependencies, etc., and be able to interpret the available data. All this poses in itself a considerable burden and cost for the claimant. Once the above requirements are fulfilled, the expert must face the causal analysis to prove fault or defect and the link with the damage. 

Last but not least, we assume that victims can claim liability against multiple liable parties, including product manufacturers, users or operators, and providers of AI systems integrated in the products. Although the burden of proof will depend on the type of defendant, our approach assumes the worst case scenario where the claimant raises the claim against one or many of the AI systems providers, needing to go into the technicalities of how the AI system works and what its state was at the time of the accident.

\subsection{Autonomous Urban Cleaning Robots}
\label{sub:cleaninrobots}

\subsubsection{Characteristics of the AI System and Operational Environment}

\begin{figure}[ht]%
\centering
\includegraphics[width=\textwidth]{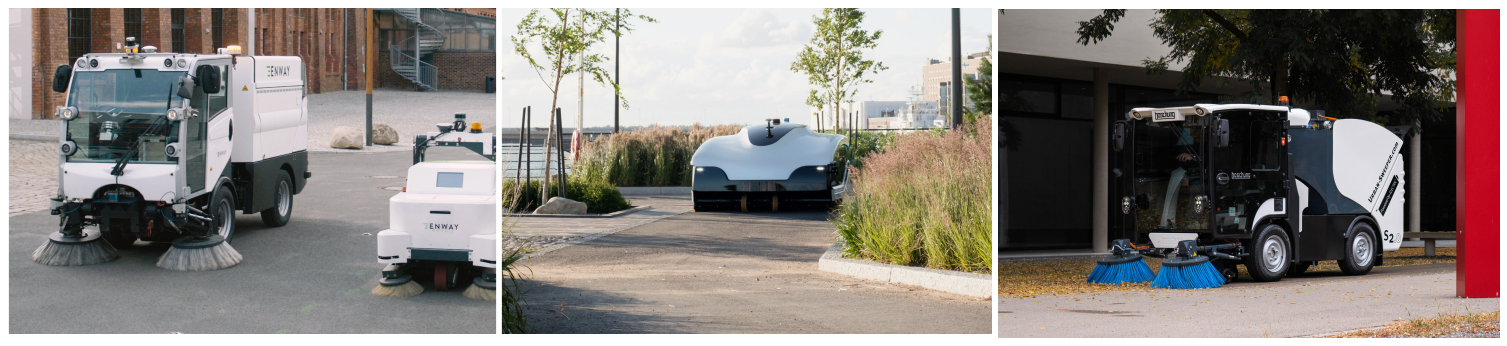}
\caption{From left to right, three examples of the current state of this kind of technology: the systems developed by \citeA{ENWAY}, \citeA{Trombia}
and \citeA{Boschung}.}\label{fig:usecase1}
\end{figure}

An autonomous fleet of cleaning robots operates in pedestrianised public areas. The robots are equipped with multiple sensors (e.g. cameras, Light Detection and Ranging (LiDAR), radar, ultrasound, GPS, etc.), digital information (e.g., digital maps), connectivity features, including communication between the robots and between the robots and the infrastructure, and actuators to perform the cleaning tasks. The robots include multiple AI/ML systems, each one responsible for a particular task (e.g. perception systems for the detection and location of litter and dirt, robot localization and mapping, detection of obstacles, trajectory planning or lateral and longitudinal control of the platform, etc.). Some examples of the current state of this kind of technology are depicted in Fig. \ref{fig:usecase1}.

Each cleaning robot belongs to a fleet deployed throughout the city. An employee is in charge of defining the operation areas to be cleaned (i.e., the missions) and monitoring multiple robots in simultaneous operation from a remote-control centre. The fleet can coordinate the safe cleaning of the selected region, the interaction with pedestrians, and the avoidance of obstacles, with a high degree of autonomy. The role of the human operator is of a supervisory nature following the human oversight strategy defined by the cleaning robot provider.

\subsubsection{Description of the Events Leading to Damage}

A colourful baby stroller is parked in front of an advertising banner with similar colour patterns while the baby's guardian looks at a nearby shop window. One of the cleaning robots does not seem to recognize the stroller as an obstacle and collides with it. The stroller is damaged and the baby slightly injured.

\subsubsection{Potential Causes of the Accident}
The accident described in this use case could have been caused by any of the following issues: 
\begin{itemize}
    \item A flaw in the vision component of the system, the AI perception model, that caused a failure in the detection of the trolley because it was somehow camouflaged with the background (due to the presence of an advertising banner with colour and pattern similar to the ones of the trolley). This led to an image segmentation error (i.e., a false negative) that considered the stroller as part of the background of the banner.  
    \item An AI cybersecurity vulnerability in the perception model that was exploited by a third party causing the AI to malfunction. For example, an adversarial machine learning vulnerability could have been exploited by vandals by placing printed stickers on the advertising banner to prevent the detection of objects around it.
    \item An attack targeting the robot's sensors, such as blinding, jamming or spoofing. In the absence of mechanisms to mitigate this type of denial-of-service attacks, the perception and planning systems can fail to detect the baby stroller, preventing the robot to avoid the collision.
    \item A cyberattack that led to the compromise of any of the robot's Electronic Control Units (ECUs). An attacker could gain unauthorised access and force the system to take an unsafe action causing the robot to collide with the stroller. The attacker could launch such an attack locally (e.g., through physical access to the robot) or remotely, either from the vicinity (e.g., through WiFi or Bluetooth access) or from Internet potentially from anywhere in the world.
\end{itemize}
On top of that, the presence of the flaw, or the conditions in which the accident took place, could have the following origins:
\begin{itemize}
    \item Failure due to an updated version of the perception system devised to reduce the number of false positives and negatives of previous versions (resulting in many regions not being cleaned correctly). The confidence threshold for considering a detection as a true obstacle was increased to reduce the number of false positives. Unfortunately, the similarity between the texture and colour of the baby stroller with the background of the advertising banner from the camera’s perspective resulted in a potential obstacle being detected with not very high confidence and discarded by the new updated version of the segmentation module.
    \item Failure of the provider of the perception system to timely distribute a software update to fix a known safety flaw or security vulnerability of the AI system. For example, if the harm had been caused due to a flaw in the segmentation mechanism, the manufacturer could have released a software update to address it by implementing a sensor fusion approach that included range-based sensor data, which would have allowed the 3D volume of the stroller to be detected as an obstacle, and be avoided by the path planning system. Similarly, if the harm had been produced as a result of the malicious exploitation of an adversarial machine learning vulnerability, the manufacturer could have released a software update to mitigate it by, for example, updating the AI model with a more resilient version trained with adversarial samples. 
    \item Failure of the operator of the system to timely apply the software update that could have been made available by the manufacturer to fix the safety flaw or security vulnerability that led to the harm described in this scenario.
    \item Failure of the remote human operator to properly supervise the operation of the robot fleet. The failure may be due to inadequate supervision by the human operator (i.e., incorrect compliance with the human oversight mechanisms defined by the provider), or to defects in the human-robot interfaces (i.e., deficiencies in the human oversight mechanisms defined by the cleaning robot producer). 

\end{itemize}

\subsubsection{Potential Liable Parties}
The autonomous cleaning robots are very complex systems with many different internal components based on a variety of AI models, which affect each other and are usually developed and integrated by different parties or subcontractors. The faulty behaviour or the defect can be in one of the components, in several components, or in a faulty integration of these components. Consequently, there could be multiple liable parties within the complex supply chain involved in the development of the cleaning robots. Therefore, potentially liable parties include:

\begin{itemize}
    \item Final producer of the cleaning robots. 
    \item Provider of individual AI components integrated in the cleaning robots (e.g. navigation, perception systems such as vision component, path planning, low-level controllers and operational interfaces). 
    \item Professional user or operator: the municipality, or a company providing the service to the municipality, deploying the cleaning robot services in the urban area.
    \item Adversaries that attack the system by exploiting vulnerabilities in the AI components (e.g., adversarial machine learning) or in the broader software and hardware surface (e.g., buffer overflows).
\end{itemize}

\subsubsection{Obtaining Compensation}
As described in the scenario, there may be multiple alternative or cumulative reasons for the damage, including low confidence detection, cyberattacks against the AI models or the underlying software and hardware architecture, etc. 
A realistic scenario is to assume that all these possible causes should be assessed by an expert opinion to prove fault or defect.
For instance, an expert could determine that the result of the perception system seems to be wrong at the time of the collision, since the stroller does not seem to appear in some list of detected objects (if available). The expert may thus be able to prove that the stroller was not properly detected (without indicating the cause). She may also be able to discard that the sensors were jammed or spoofed since the raw data seems correct (raw data should be interpretable). The expert could further suppose a correlation between such detection failure and the control decision of the robot to move forward until colliding with the stroller. This may allow the claimant to establish \emph{prima facie} evidence. However, proving correlation does not allow discarding alternative causes of the damage (e.g., the stroller could have moved towards the robot abruptly because it was unattended without a brake). 

With regard to the lack of causality and opacity features of the AI systems, it may be impossible to infer a clear causal link between a specific input and the harmful output. Concerning the unpredictability feature, it is possible that the same robot, in an almost identical scenario, but with slight variations in lighting, would have behaved correctly (which could be used as evidence in favour of the defendant). As for self and continuous learning, we assume that type-approval procedures would not allow such an approach while the cleaning robots are in operation.

It is worth noting that, in any case, the expert would require access to the the logs of the cleaning robot and technical information about the internal AI systems in order to be able to conduct a forensic analysis of the incident and reconstruct the facts that led the robot behave in the way it did. This type of information is often only available to the manufacturer of the robot.

\subsection{Autonomous Delivery Drones}
\label{subsec:drones}

\subsubsection{Characteristics of the AI System and Operational Environment}
A fleet of autonomous delivery drones (a.k.a. unmanned aerial vehicles or unmanned aircraft) is used to transport small and medium-sized packages (maximum 15 kg) to people's homes in rural and suburban areas, where there are sufficiently large landing areas (of at least 2m$^2$ due to the size of the drones) to release cargo safely and without interaction with end-users. The drone is a multirotor quadcopter with horizontally aligned propellers. The drones are autonomous in the sense that the operator loads the cargo into the drone, establishes the final destination, and the drone is capable of performing autonomous vertical take-off and landing, navigating to destination, dropping off the parcel and returning to origin without the need for a pre-defined key points route. They are capable of detecting and avoiding possible obstacles within the planned local route (e.g., birds, high voltage lines, trees, etc.), being robust against moderate wind disturbances. 
The drones are equipped with multiple sensors and communication systems. Inertial Measurement Units (IMUs) are used to calculate the orientation, altimetric pressure, velocity, rotation rate, angular velocity and tilt, linear motion and heading of the drone. GNSS is used to perform global localization and navigation. LiDAR sensors and digital cameras are used as the input to sense the environment, detect obstacles, and ensure a clear and safe landing zone. The drones have short-range communication systems within the visual line of sight (VLO) with the operator, but in order to continue monitoring the system beyond the visual line of sight (BVLO), they also have 3G/4G/5G cellular communications.

The drones make use of multiple AI components, each one responsible for a particular task, e.g., scene understanding and obstacle detection, autonomous localization and navigation, etc. They include four main operation modes: (1) global planning, (2) take-off, (3) global and local navigation and (4) landing. The most critical tasks are vertical take-off and landing, where the drone continuously monitors that the planned local trajectory is clear of any obstacles. Depending on the weight of the cargo and wind conditions, the delivery mechanism may involve landing the drone or landing the delivery by hovering a few meters above the delivery spot while releasing a wire with the delivery tethered to it until it reaches the ground. 

The operator must comply with all requirements established by applicable civil aviation regulations including the verification that the lighting and weather conditions for the intended trip (outbound and return) are within the specifications provided by the provider. They must check weight and dimensions of the cargo, the battery status and available range, and verify that the landing conditions at the destination are as required by the provider’s specifications. Only if all required conditions are met, the drone can be loaded with the cargo and launched for delivery. Thanks to the mobile communication interface, the operator can monitor the entire process remotely. 

\begin{figure}[ht]%
\centering
\includegraphics[width=\textwidth]{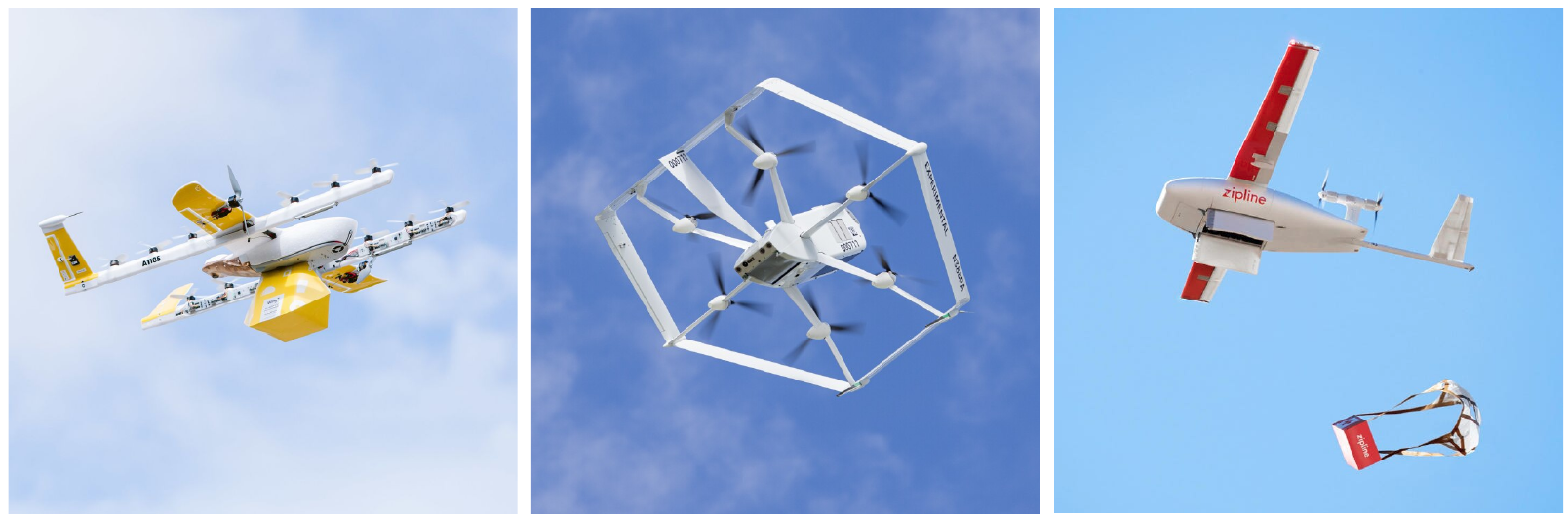}
\caption{From left to right, three examples of the current state of this kind of technology: the systems developed by \citeA{Wing}, \citeA{PrimeAir}
and \citeA{Zipline}.}\label{fig:usecase2}
\end{figure}

Some examples of current developments in autonomous delivery drones technology are depicted in Fig. \ref{fig:usecase2}.

\subsubsection{Description of the Events Leading to Damage}
The drone is approaching a suburban area to deliver a cargo. The day is a bit windy and cloudy, so the delivery mechanism selected by the operator was by landing the drone. Once the delivery point has been detected and checked that it is clear of obstacles, the drone starts the vertical landing manoeuvre. 

A child from the neighbouring house is playing in his garden and, when he hears the drone approaching, he suddenly runs towards it. The vertical landing is not properly aborted, and the drone eventually hits the child. As a result of the impact, the drone's control system becomes unstable and the drone crashes into a nearby car parked in the street. The child is seriously injured, and the car undergoes considerable damage.

\subsubsection{Potential Causes of the Accident}
The accident could have been caused by any of the following issues:

\begin{itemize}
    \item A flaw in the AI-based obstacle detection system that caused a failure in the detection of the child approaching to the landing zone. This failure could have been produced due to multiple reasons. For example, some bias or mislabelling in the training data of the perception system related to small objects on the ground or inadequate lighting conditions.
    \item A flaw in the AI-based decision-making and control systems of the drone that caused it not to react in time to the sudden presence of the child, properly correcting the trajectory. This could have been due to several reasons, such as insufficient reaction time of the control system or because of stability problems due to inappropriate wind conditions.
    \item Failure of the AI systems to handle evolving unfavourable weather conditions. The lighting and weather conditions were adequate at the time the operator made the decision to deliver the parcel with the autonomous drone but worsened as the drone approached its destination. No self-diagnosis functions were incorporated into the autonomous delivery drone and the mission was not remotely aborted by the operator. Poor lighting conditions would have reduced the accuracy of the perception systems. Inappropriate wind conditions would have compromised the control system's ability to perform obstacle avoidance manoeuvres. 
    \item A deliberate cyberattack on the drone’s systems, targeting the drone's sensors (e.g., sensors blinding, jamming or spoofing), exploiting an AI cybersecurity vulnerability (e.g., an adversarial machine learning attack to the camera-based perception system) or exploiting a cybersecurity vulnerability in the broader hardware and software architecture (e.g., unauthorised  access to the internal systems of the drone through the wireless interface or cellular communications).
\end{itemize}

\subsubsection{Potential Liable Parties}
Similarly to the cleaning robots described in Section \ref{sub:cleaninrobots}, autonomous delivery drones are very sophisticated systems with many different internal components, some of them making use of different kind of AI models. These systems are designed to interact with each other, and can be developed and integrated by different parties and subcontractors. 

In this context, the origin of the faulty behaviour or defect can be in one of component, in several components, or in their specific integration. Therefore, there are multiple potentially liable parties, including:

\begin{itemize}
    \item Final producer or manufacturer of the autonomous delivery drones.
    \item Provider or manufacturer of individual AI components integrated in the drone (e.g., localization, navigation, perception systems, low-level controllers, take-off, landing and delivery mechanisms, and operational interfaces).
    \item Professional user or operator: the company providing the delivery service deploying the autonomous delivery drones in the rural and suburban areas.
    \item Adversaries that attack the system by exploiting vulnerabilities in the AI components (e.g. adversarial machine learning) or in the broader software and hardware surface (e.g. jamming or spoofing of sensor signals, or buffer overflow vulnerabilities in the software implementation).
\end{itemize}

\subsubsection{Obtaining Compensation}

Let us consider that an expert has access to the inputs/outputs and internal states of the perception system a few minutes before the accident and during the accident, as well as the technical documentation to enable 
its correct interpretation. One of the first hypotheses to investigate would be the case that the child was not correctly detected by the perception module. For example, if the system records some list of moving obstacles detected on the ground, the expert could check whether or not any obstacle could be associated with the boy on that list. If no obstacles were detected before or during the accident, this could be presented as evidence of fault or defect, and may also serve to presume a causal link to the damage. This is the best case scenario, but this internal list may not be accessible or available. The perception system could provide other types of information represented in a format not directly interpretable by humans (linked with the lack of causality and opacity features), as input to the decision-making, path planning and control modules. 

If it is possible to demonstrate that the child was properly detected (e.g., with the aforementioned list of obstacles), then the expert would have to investigate why the drone’s decision making, path planning and control systems did not avoid the collision. Some internal information would be needed on environmental conditions measured by or communicated to the drone (e.g., wind speed, lighting conditions), drone status (e.g., height, rotors speeds, pose), intended route (e.g., the local trajectory of the drone) and control actions (e.g., target and current speed of all rotors). If all this information were available, it would even be possible to reconstruct the accident and verify that the planned route and control actions did not prevent it. This could be presented as evidence of fault or defect, and link to the damage. Again, this is a favourable scenario, but this information may not be directly available, or not directly interpretable (i.e., opacity). 

It could be the case that the entire system is based on a completely opaque end-to-end model (from the raw sensor inputs to the final control actions on the rotors, without learning causal relationships), so that intermediate representations are neither available nor interpretable. Under these circumstances, it is possible that an expert may be able to establish some correlation between some possible alternative causes and the damage caused to the child and the parked vehicle. However, the lack of causality and the opacity of the the drone’s AI systems can make it impossible to establish a clear causal link between any of the possible alternative causes and the accident. In addition, if an analysis of raw sensor data (e.g., IMUs, cameras, LiDAR, GNSS) shows that the sensors were jammed or spoofed, it would be very difficult to determine the source of the attack as sensor data would be compromised. As for the unpredictability feature, an added difficulty is that the defendant may be able to prove that the same delivery drone, in an almost identical scenario, but with slight variations in weather conditions, would have been able to safely abort the delivery manoeuvre. Finally, regarding self and continuous learning, we can reasonably assume that the regulations for type-approval of these products would not allow their implementation once the systems are deployed and operational.

\subsection{Robots in Education}  

\subsubsection{Characteristics of the AI Systems and Operational Environment}

Socially assistive robots (a.k.a. social robots) typically are used in complex physical environments with the purpose to interact with humans \cite{dautenhahn2007socially}. 
They can integrate into the social environment and, autonomously or with limited human supervision, navigate physical space and interact with humans, usually in restricted settings such as healthcare environments \cite{Fosch-Villaronga2022}.
They usually have explicit or implicit anthropomorphic features and they can perform across a spectrum of behavioural policies which typically depend on their morphology. 

\begin{figure}[ht]%
\centering
\includegraphics[width=\textwidth]{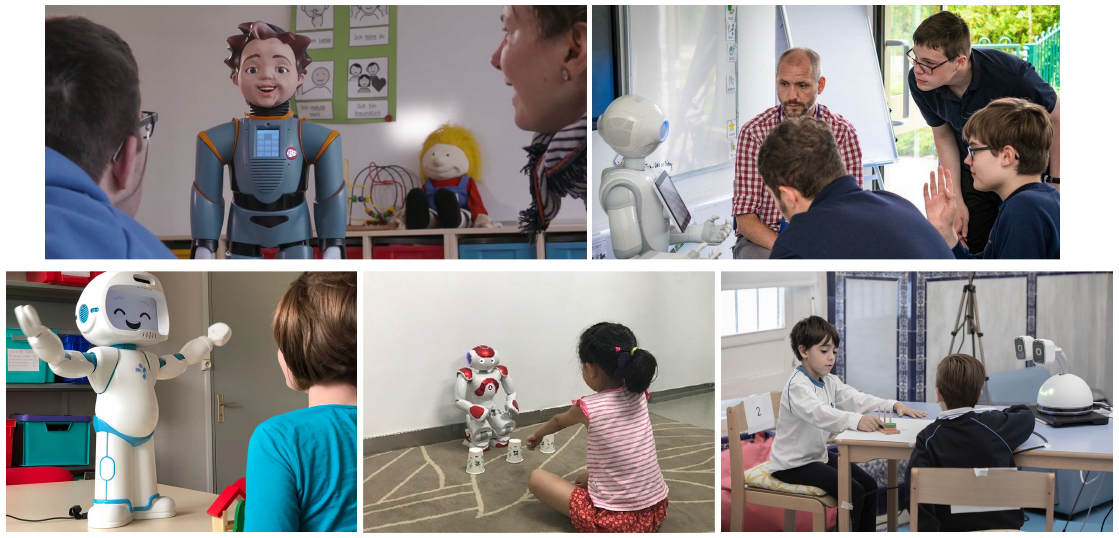}
\caption{From left-to-right, top-to-bottom, five different robotic platforms in the context of education: 
\citeA{De-Enigma}, Pepper \cite{Pepper}, QTrobot from LuxAI \cite{LuxAI}, Nao robot \cite{Nao2019} and Haru \cite{Charisi2020}.}\label{fig:usecase3}
\end{figure}

Despite the diverse characteristics of robots, for the present use case we consider a robot that has a configuration that would allow its effective use in educational contexts to support the socio-emotional skills of autistic children. In this sense, the robot is mobile and includes perception components,  navigation, facial, speech and emotion recognition, localization, decision-making, mapping and path planning systems, manipulation, grasping, expressive communication and other AI-based systems. It is about 1.30m tall and it has arms, mainly for gesturing, as an expressive behaviour. The robot is equipped with multiple sensors to detect the environment, including cameras, 3D sensors, laser, sonar infrared, tactile, microphones and inertial measurements units. It is equipped with a tablet as an alternative means for communication. It can perceive and process speech using AI systems, including a module for verbal communication with expressive voice.  It is capable of detecting obstacles, people, and facial expressions using AI-based computer vision algorithms.  Lastly, it is equipped with an AI-based cognitive architecture which combines task-related actions and socially adaptive behaviour for effective and sustained human-robot interaction. Some examples of similar prototypes already tested in operational environments are depicted in Fig. \ref{fig:usecase3}.

In addition, for the robot to be tailored for that specific application (autistic children in school environments) it comes with pre-designed interventions for cognitive engagement (task-oriented). The robot is capable to adapt its social behaviour according to the child's personal abilities and preferences in the context of autism. 

Focusing on the need for inclusion and preparation of autistic children for an independent life, a mainstream primary school school unit decided to increase the number of admissions for high functioning autistic children. However, the staff is not adequate to undertake individual support for the autistic children. For this reason, the director of the school decides to introduce one robot per class for personalized interaction with the autistic children for the improvement of their social skills. The school signs a contract with the company providing robotic educational services using the robot (as well as all integrated AI systems) to provide the educational services at the school.  

The robots are installed in the classrooms for regular personalized interventions with the autistic children and for voluntary interaction during children’s free time. The robots are mobile, and they can navigate dedicated space during the children’s free time if a child requests so. The robot learns from the interaction with the autistic children and adapts its social behaviour. While during the lesson time the robot is available only for children with autism to perform personalized interventions, during the free time, any child of the school can interact with the robot at dedicated spaces. 

\subsubsection{Description of the Events Leading to Damage}

In this use-case, we focus on harm which might be caused because of the adaptive behaviour of the robot. Some property damage may also occur. We propose three different scenarios. 

\textbf{Scenario 1}: Physical harm and property damage towards a child with darker skin. 
The robot fails to perceive a child with darker skin, and it causes physical harm to the child. The blow caused by the robot also resulted in the breakage of the child's glasses. 

\textbf{Scenario 2}: Physical harm and property damage towards a child that behaves in an unexpected way. 
The robot fails to respond in an appropriate way to an autistic child that might have unexpected behaviour and hits the child. The blow caused by the robot also resulted in the breakage of the child's glasses. 

\textbf{Scenario 3}: Long-term psychological harm towards a neurotypical child. During children's free time at the school, a neurotypical child interacts with the robot on a regular basis. The robot adapts to the child's needs and requests which subsequently leads the child to develop potentially medically recognised pathological such as addictive behaviour towards the robot (e.g., increase preference of the child to interact with the robot rather than with humans causes an abnormal socio-emotional development), depression (e.g., social isolation can negatively influence psychological health leading to depressive symptoms), emotional distress (e.g., the inappropriate robot response in scenario 2 leads to psychological trauma), or abnormal cognitive and socio-emotional development and dependencies.

\subsubsection{Potential Causes of the Accident}

For the scenarios 1 and 2 
the damage could have been caused by the following issues:
\begin{itemize}
    \item Flaw in the robot's perception module that does not perceive the child due to biases in the robot's perception system with respect to children with certain physical characteristics. 

    \item Flaw on the decision-making and path planning modules of the robot which fails to adapt to the child user, from a prolonged period of interaction with children with certain behavioural characteristics.

    \item Flaw on the control module which fails to consider the physical and behavioural differences of the child user. 
\end{itemize}

For the scenario 3 the damage could have been caused by the following issue: 

\begin{itemize}
    \item Robot adaptation: the adaptation module of the robot embeds an intrinsic motivation element which contributes to the human-robot sustained interaction. The robot's internal goal to remain in an optimal level of comfort for the child-user contributes to its adaptation to the specific child's characteristics, needs and behaviours. This robot behaviour develops a closed loop of cognitive and socio-emotional interaction with the child that might lead to the child's addiction to the specific robot behaviour. In a long-term interaction the child might exhibit a preference for interaction with the robot rather than human social agents. In that case, the child and the robot develop in a mutual adaptation loop.  
\end{itemize}

\subsubsection{Potential Liable Parties}
Social robots can be very complex systems with many different AI-based components that are integrated in a single platform. These components fall into three main categories (i) perception, decision-making and planning, and action and execution, and they need to interact with each other. Similar to the use cases presented in Sections \ref{sub:cleaninrobots} and \ref{subsec:drones}, the origin of the faulty behaviour or defect can be in one of the components, in multiple components or in their integration. In addition, for the specific scenario, there might be modules that refer to a task-specific robot behaviour and other modules that relate to the robot's social interaction. For the task-specific robot behaviour, a separate company might be involved who are specialists in pedagogy and autistic children.

As such, there are multiple potentially liable parties including:

\begin{itemize}
\item Manufacturer of the robot who is also the provider of the robot's AI systems.
\item  Providers of the AI modules integrated into the robot before it is placed on the market.
\item Provider of the educational system which provides the task-specific modules.
\item Company using the robot to provide educational services.
\item The school that makes use of educational robotic services. 
\end{itemize}

\subsubsection{Obtaining Compensation}

For product liability the victim should prove defectiveness of the robot and link with the damage. For fault-based liability, the victim should prove negligence by some of the potentially liable parties and link with the damage. As described above the compensation claims can be directed against the robot manufacturer, the providers of some of the AI components, or the user (the company providing the educational services). Although the victim could also claim against the school, it should be noted that the robot was intended to function without supervision by a teacher, and therefore courts are unlikely to uphold such a claim. 

On the one hand, courts may or may not infer defect or fault and causality from the fact that the robot caused the relevant injuries. The fact that the AI-systems influencing the robot’s behaviour adapted during the latter’s autonomous operation (e.g., by means of self and continuous learning) may put into doubt such an inference. Courts may namely take into account that the robot’s behaviour depends on various circumstances that may be considered unforeseeable for the provider or user (namely the precise operating environment, human interaction and the input data the robot is exposed to). Demonstrating the extent to which the robot's mechanism of adaptation to the behaviour of the children it interacts with led to a change in behaviour not foreseen by the provider (resulting in harm) is quite complex. On the victim's side, it would require, at the very least, expert knowledge and access to a considerable level of documentation regarding both the system design and the logs while it was in operation.  

On the other hand, regarding potential issues not linked with self and continuous learning, proving defect or negligence, and causal link with the damage, would require expert analysis of the design and functioning of the robot, the relevant AI-systems, or the human oversight mechanisms foreseen for users. In addition, the injured parties should be entitled to access the necessary level of  information for the purposes of their claim, including the aforementioned logs while the robot was in operation. On the basis of such information, an expert may notably be able to determine whether the result of the robot’s perception system is correct at the time of the accident, for instance by checking whether the physically injured child appears in the list of detected objects. The expert may also review the relevant control decisions of the robot, e.g. the decisions to interact in a certain way with the affected children, or the decision to actuate certain movements. The analysis may also inform the supposition of a correlation between, for instance, a detection failure and a relevant control decision of the robot, but establishing a clear causal link will be very difficult due to the aforementioned specific AI features. Regarding the role of the user, with a sufficient level of access to information, such as the user oversight requirements (e.g., instruction of use), and data logged during the operation, the victim could establish a possible failure to comply with the user’s standard of safety or care. All these elements could serve as evidence to prove defect or negligence. However, due to the complexity of the system as a whole, as well as the lack of causality, unpredictability or opacity of AI systems, it would be very complex, even for an expert, to establish clear causation between the specific operating conditions and the harmful event. For example, it would be quite complex to exclude other elements such as inappropriate behaviour on the part of children or supervisors. 

\section{Conclusions}\label{conclusion}

In this work, we presented three hypothetical use cases of products driven by multiple AI systems, operating autonomously in non-private environments, causing physical harm (including one scenario involving mental health) and property damage to third parties. These use cases, selected based on a set of  inclusion criteria, represent AI technologies with a high level of readiness operating in real-world scenarios. We described them realistically and with a high degree of technical granularity. In addition, we explored the scenarios from the perspective of product and fault liability regimes, identifying the possible causes of the accident, the liable parties, and describing the technicalities underlying the process of obtaining compensation from the victim. 

Through this process, we highlighted the technical difficulties that an expert opinion would face in trying to prove defect or negligence, and the causal link to the damage. This is due to certain characteristics that some AI systems intrinsically contain, namely lack of causality, opacity, unpredictability and self and continuous learning  We attempted to provide certain degree of complementarity between the three use cases to allow the analysis of different factors when dealing with the burden of proof. As a further contribution, we identified and described these specific AI features in detail, so that they can serve as a basis for other studies related to legal issues and AI. Our analysis indicates that liability regimes should be revised to alleviate the burden of proof on victims in cases involving AI technologies. How to most effectively adapt liability regimes to meet the new challenges specific to AI systems is still an open question for further research. In addition, future works can also be directed towards the development of a formal model to guide possible arguments in liability cases involving AI systems.

\acks{The authors acknowledge funding from the HUMAINT project at the Digital Economy Unit at the Directorate-General Joint Research Centre (JRC) of the European Commission. The authors also acknowledge the valuable contributions of Mr. Bernd Bertelmann and Ms. Ioana Mazilescu.}

\vspace{0.5cm}
\textbf{Disclaimer:} The views expressed in this article are purely those of the authors and may not, under any circumstances, be regarded as an official position of the European Commission. 

\vskip 0.2in
\bibliography{sample}
\bibliographystyle{theapa}

\end{document}